\pdfoutput=1

\documentclass[runningheads]{llncs}

\usepackage[T1]{fontenc}
\usepackage[utf8]{inputenc}

\usepackage{graphicx}

\usepackage{hyperref}
\usepackage{color}

\usepackage{siunitx}
\let\model\textsc
\newcommand{\copybs}[0]{\model{copy}}

\newcommand{\lstm}[0]{\model{LSTM}}
\newcommand{\trmmodel}[0]{\model{TRM}}

\newcommand{\trmono}[0]{\model{mono}}
\newcommand{\trmulti}[0]{\model{multi}}

\newcommand{\trmultiC}[0]{\trmulti{}}

\newcommand{\uacc}[0]{type accuracy}

\newcommand{\sig}[0]{SIGMORPHON}
\newcommand{\tr}[0]{Transformer}

\newcommand{\ltf}[0]{lemma-tag-form}

\newcommand{\ud}[0]{Universal Dependencies}

\newcommand{\en}[0]{English}

\newcommand{\ie}[0]{Indo-European}

\newcommand{\writtenLatinScript}[0]{written in the Latin script}

\newcommand{\pb}[1]{\bfseries #1\normalfont{\textsuperscript{*}}}

\newcommand{\multiletterM}[0]{\tau}

\newcommand{\paramsMono}[0]{3.12M-3.69M}
\newcommand{\params}[0]{5.69M}

\newcommand{\citeSTs}[0]{\cite{st18-cotterell-etal-2018-conll,st17-cotterell-etal-2017-conll,st16-cotterell-etal-2016-sigmorphon,st23-goldman-etal-2023-sigmorphon,st22-kodner-etal-2022-sigmorphon,st21-pimentel-ryskina-etal-2021-sigmorphon,st20-vylomova-etal-2020-sigmorphon}}

\newcommand{\repo}[0]{\url{https://github.com/tomsouri/multilingual-inflection}}

\usepackage[textsize=footnotesize, backgroundcolor=yellow!25, linecolor=black!25]{todonotes}

\newcommand{\tom}[1]{\todo[backgroundcolor=orange!50]{\textit{\footnotesize Tomas}: #1}}
\newcommand{\tomH}[1]{\todo[backgroundcolor=violet!25]{\textit{\footnotesize Hidden note}}}
\newcommand{\ftodo}[1]{\todo[backgroundcolor=yellow]{\textit{\footnotesize Future}: #1}}
\newcommand{\ales}[1]{\todo[backgroundcolor=green]{\textit{\footnotesize Aleš}: #1}}

\renewcommand{\tom}[1]{}
\renewcommand{\tomH}[1]{}
\renewcommand{\ftodo}[1]{}

\renewcommand{\ales}[1]{}

\usepackage{tikz}
\usetikzlibrary{positioning} 
\usetikzlibrary{arrows.meta}
\usetikzlibrary{decorations.pathreplacing}

\usepackage{booktabs}

\usepackage[absolute]{textpos}

\begin{document}

\begin{textblock}{16}(0,0.1)\centerline{This paper was published in \textbf{Text, Speech, and Dialogue (TSD) 2025}}\end{textblock}
\begin{textblock}{16}(0,0.3)\centerline{-- please cite the published version {\small\url{https://doi.org/10.1007/978-3-032-02551-7_5}}.}\end{textblock}

\title{Flexing in 73 Languages: A Single Small Model for Multilingual Inflection}

\author{Tomáš Sourada\orcidID{0009-0003-6792-825X} \and
Jana Straková\orcidID{0000-0003-0075-2408}}

\institute{Charles University, Faculty of Mathematics and Physics,\\Institute of Formal and Applied Linguistics, Prague, Czech Republic \\
\email{\{sourada,strakova\}@ufal.mff.cuni.cz}
}

\maketitle              

\begin{abstract}

We present a compact, single-model approach to multilingual inflection, the task of generating inflected word forms from base lemmas to express grammatical categories. Our model, trained jointly on data from 73 languages, is lightweight, robust to unseen words, and outperforms monolingual baselines in most languages. This demonstrates the effectiveness of multilingual modeling for inflection and highlights its practical benefits: simplifying deployment by eliminating the need to manage and retrain dozens of separate monolingual models.

In addition to the standard \sig{} shared task benchmarks, we evaluate our monolingual and multilingual models on 73 Universal Dependencies (UD) treebanks, extracting \ltf{} triples and their frequency counts. To ensure realistic data splits, we introduce a novel frequency-weighted, lemma-disjoint train-dev-test resampling procedure.

Our work addresses the lack of an open-source, general-purpose, multilingual morphological inflection system capable of handling unseen words across a wide range of languages, including Czech. All code is publicly released at: \repo{}.

\end{abstract}

\section{Introduction}

\begin{figure}

\begin{center}

\resizebox{1\hsize}{!}{

\tikzstyle{rec}=[rectangle, rounded corners=1ex, minimum height = 0.8cm, text height=1.5ex, text depth=0.5ex]

\begin{tikzpicture}[thin, inner sep=0.7ex]

\node[rec, draw] (a) {w};
\node[rec, draw, right=0cm of a] (b) {e};
\node[rec, draw, right=0cm of b] (c) {l};
\node[rec, draw, right=0cm of c] (d) {l};
\node[rec, draw, right=0cm of d] (e) {UPOS=ADV};
\node[rec, draw, right=0cm of e] (f) {Degree=Sup};

\node[rec, draw, right=2cm of f] (g) {b};
\node[rec, draw, right=0cm of g] (h) {e};
\node[rec, draw, right=0cm of h] (i) {s};
\node[rec, draw, right=0cm of i] (j) {t};

\draw [thick, -{Latex[length=2mm]}] (f) -- (g) node[midway, above] {Inflection};
\draw [thick, -{Latex[length=2mm]}] (f) -- (g) node[midway, below] (arr) {system};

\draw[decorate,decoration={brace,amplitude=5pt, mirror}] (a.south west) -- (d.south east) node[midway, below=6pt] {lemma};

\draw[decorate,decoration={brace,amplitude=7pt, mirror}] (e.south west) -- (f.south east) node[midway, below=8pt] {morphological tags};

\draw[decorate,decoration={brace,amplitude=5pt, mirror}] (g.south west) -- (j.south east) node[midway, below=8pt] {form};

\end{tikzpicture}
}
\end{center}
\caption[The Inflection Task: input-output example]{The morphological inflection task: an input-output example, inflection of \en{} lemma ``well'' to superlative, inflected form ``best''.} 
\label{fig:inflection-task-input-output-formulation}
\end{figure}
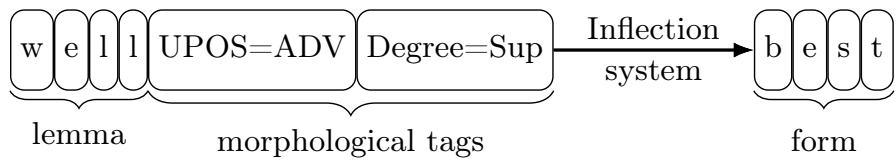

Morphological inflection, the process of modifying a base word form (lemma) to express grammatical categories (see example in Fig.~\ref{fig:inflection-task-input-output-formulation}), is a well-established and practically significant task.

Despite the field's high level of activity, driven mainly by the \sig{} shared tasks \citeSTs{},
there is still no open-source, lightweight, neural morphological inflection generator for unknown (out-of-vocabulary, OOV) words in Czech, let alone one supporting multiple languages, available within the Czech academic environment.
MorphoDiTa~\cite{morphodita-strakova-2014-open} relies on a morphological dictionary for Czech and lacks an out-of-vocabulary (OOV) guesser for generation.\footnote{Although MorphoDiTa includes OOV heuristics for morphological analysis, the inverse process of generation, it does not support OOV generation.} UDPipe \cite{udpipe:2017} does not support inflection generation at all. And although Sourada et al. (2024) \cite{oovs-sourada-2024-lrec} provide an OOV guesser, it is limited to Czech nouns.

We address these shortcomings by experimenting with supervised deep learning models for morphological inflection, based on the \tr{} architecture,
with a focus on potential future production deployment and multilingual applicability.

Our multilingual system, trained jointly on 73 languages, is extremely lightweight, consisting of a single model with \params{} trainable parameters, capable of performing inflection across all 73 languages. The model handles all parts of speech, inflects unknown words (with evaluation conducted entirely on words unseen during training), and crucially, outperforms both the baseline systems and separately trained monolingual models in most languages. Furthermore, it offers practical advantages for deployment: a single multilingual model eliminates the need to manage dozens of individual models in memory and significantly simplifies maintenance, avoiding the overhead associated with retraining 73 separate models.

We make all our work open-source by publicly releasing our code at GitHub: \repo{}.

\section{Related Work}

Morphological inflection has seen considerable research interest in the recent years, mainly thanks to the \sig{} shared tasks held from 2016 to 2023 \citeSTs{}.
This period was marked by major advancements in model architectures, particularly recurrent neural networks and \tr{}s.

The field has also undergone significant methodological changes, such as the shift from na\"ive splitting by form to a \textit{lemma-disjoint} split \cite{goldman-2022-lemma-overlap}, and has identified weaknesses in previously established practices. Notable examples include the deficiencies of the original uniform train-dev-test split, as compared to the \textit{frequency-weighted} split proposed by Kodner et al. (2023) \cite{kodner-reality-2023} to achieve a more realistic train-test distribution. However, these challenges have not yet been fully addressed.

Furthermore, many previous studies were experimental and/or trained on small datasets, making the transition from academic research to production deployment challenging.

\section{Data}

We use the \sig{} 2022 \cite{st22-kodner-etal-2022-sigmorphon} and the \sig{} 2023 \cite{st23-goldman-etal-2023-sigmorphon} shared task benchmarks for evaluating the strength of our models.\footnote{Hebrew was excluded from SIGMORPHON 2023 due to filename inconsistencies that prevented reliable comparison.}
Furthermore, with the goal of future deployment and multilingual applicability, we select 73 languages
of the \ud{} \cite{ud2-nivre-etal-2020} corpora\footnote{The selection contains 32 \ie{} languages \writtenLatinScript{} and 41 languages without restriction on language family or script. For languages with multiple corpora in UD, we chose one corpus per language based on historical preferences, practical considerations, and the defaults in the widely adopted UDPipe tool \cite{straka-2018-udpipe}. We prioritized large, canonical corpora, excluded those with narrow or specialized domains, and favored manually over semi-automatically annotated corpora. Corpora lacking lemma annotations were excluded.} and lexicalize them by extracting unique \ltf{} triples along with their occurrence counts for training and evaluation.\footnote{Unlike UniMorph, the Universal Dependencies data can be more prone to noise. We rely on a neural network to mitigate its impact.}

\section{Methodology}

\subsubsection{Data Splitting}

Strategies for data splitting in the inflection field differ in the degree of overlap control between train-test splits, from an uncontrolled overlap (where \ltf{} may overlap randomly) to varying degrees of control over lemma and/or morphological feature overlap. The former has been the traditional approach for decades but has recently come under critique for hindering evaluation of models' generalization abilities.

At the very least, a \textit{lemma-disjoint} option
\cite{goldman-2022-lemma-overlap} is recently advisable, and \sig{} 2022 \cite{st22-kodner-etal-2022-sigmorphon} probed this setting in its \textit{feature overlap} setting: A test pair's feature set is attested in training, but its lemma is novel. The latest \sig{} shared task installation completely moved to the \textit{lemma-disjoint} setting in 2023 \cite{st23-goldman-etal-2023-sigmorphon}. We use the official train-dev-test split for both the \sig{} benchmarks.

Another axis for consideration is the manipulation with the \ltf{} corpus frequencies during sampling into the train-test splits. Traditionally, the corpus frequencies are neglected (or not even present in the mostly used data source, UniMorph \cite{unimorph-20-kirov-2018}) and the \ltf{} triples are sampled uniformly. Kodner et al. (2023) \cite{kodner-reality-2023} argued that uniform sampling leads to an unrealistic train-test split, with train set unnaturally biased towards low-frequency types. They introduced a \textit{frequency-weighted sampling} method to produce a more realistic train-test distribution.
However, the suggested split is not lemma-disjoint. Also, no further work has experimented with that split.

Due to the lack of standardized practice for sampling with corpus frequencies, we design
a new train-dev-test split technique for the UD datasets that satisfies both recent methodological standards (being \textit{lemma-disjoint} \cite{st23-goldman-etal-2023-sigmorphon,goldman-2022-lemma-overlap}) and practical requirements for real-world deployment by ensuring a realistic distribution of items with varying corpus frequencies through \textit{frequency-weighted sampling} \cite{kodner-reality-2023}.

\subsubsection{Model}

We employ a \tr{} architecture of the encoder-decoder, sequence-to-sequence type, characterized by a small capacity
according to state-of-the-art standards in the inflection field.
This model is trained from scratch on inflection data, using characters as tokens, with the input consisting of lemma-tag pair and the output being the corresponding inflected form (see Fig.~\ref{fig:inflection-task-input-output-formulation}).

\begin{table*}[t]
    
\centering

\caption[\sig{} 2022 comparison]{\sig{} 2022 (Feature Overlap): 
Comparison across all 16 development languages that provided both large training dataset (7k examples) and test data under the feature overlap evaluation condition.
Reported exact-match accuracy (\%). 
\ddag{}~marks datasets with less than 100 test items. The systems are: CLUZH \cite{wehrli-etal-2022-cluzh}, Flexica \cite{sherbakov-vylomova-2022-flexica}, OSU \cite{elsner-court-2022-osu}, TüM \cite{merzhevich-etal-2022-tum}, UBC \cite{ubc-yang-etal-2022-generalizing}, and baselines \cite{st22-kodner-etal-2022-sigmorphon}, Sou is Sourada et al. (2024) \cite{oovs-sourada-2024-lrec}.}
\label{tab:sig22-comparison}

\begin{tabular}{cccccccccccc} 

\toprule
& \multicolumn{5}{c}{Submitted systems} & \multicolumn{2}{c}{Baselines} & \multicolumn{2}{c}{Sou} & \multicolumn{2}{c}{OURS}\\
\cmidrule(lr){1-1} \cmidrule(lr){2-6} \cmidrule(lr){7-8} \cmidrule(lr){9-10} \cmidrule(lr){11-12}

lang & CLUZH & Flexica & OSU & TüM & UBC & Neural & NonNeur & \lstm{} & \trmmodel{} & \trmono{} & \trmultiC{} \\

\cmidrule(lr){1-1} \cmidrule(lr){2-6} \cmidrule(lr){7-8} \cmidrule(lr){9-10} \cmidrule(lr){11-12}
ang & 76.6 & 64.4 & 73.7 & 71.9 & 74.1 & 73.4 & 68.7 & 76.3 & 75.5 & 73.4 & \textbf{77.3} \\
ara & 81.7 & 65.5 & 78.7 & 78.5 & 65.5 & 81.9 & 50.8 & 79.2 & \textbf{82.6} & 81.0 & 79.6 \\

asm$^{\ddag}$ & 83.3 & 75.0 & 75.0 & \textbf{91.7} & 83.3 & 83.3 & 83.3 & 83.3 & 83.3 & 83.3 & 83.3 \\
got & 92.9 & 41.4 & \textbf{94.1} & 91.7 & 91.7 & 93.5 & 87.6 & 92.3 & 92.3 & 91.1 & 91.7 \\
hun & 93.5 & 62.9 & 93.1 & 92.8 & 91.5 & \textbf{94.4} & 73.1 & 92.8 & \textbf{94.4} & 93.2 & 93.2 \\
kat & 96.7 & 95.7 & 96.7 & 96.7 & 96.7 & 97.3 & 96.7 & 97.3 & \textbf{97.8} & 96.7 & 97.3 \\
khk$^{\ddag}$ & 94.1 & 47.1 & 94.1 & 94.1 & 88.2 & 94.1 & 88.2 & \textbf{100.0} & 94.1 & 94.1 & 94.1 \\
kor$^{\ddag}$ & \textbf{71.1} & 55.4 & 50.6 & 56.6 & 60.2 & 62.7 & 59.0 & 49.4 & 62.7 & 63.9 & 62.7 \\
krl & 87.5 & 69.8 & 85.9 & 57.8 & 85.4 & 57.8 & 20.8 & \textbf{89.1} & 85.9 & 85.9 & 86.5 \\
lud & 87.3 & 92.0 & 92.9 & 93.4 & 88.2 & \textbf{94.3} & 93.4 & 89.2 & 92.0 & 91.5 & 93.4 \\
non$^{\ddag}$ & 85.2 & 77.0 & 85.2 & 80.3 & \textbf{90.2} & 88.5 & 80.3 & 83.6 & 88.5 & \textbf{90.2} & 86.9 \\
pol & \textbf{96.1} & 85.9 & 94.9 & 74.0 & 95.7 & 74.4 & 86.3 & \textbf{96.1} & 95.6 & 95.7 & 95.5 \\
poma & 76.1 & 54.5 & 70.1 & 69.4 & 73.3 & 74.1 & 47.8 & 75.2 & \textbf{76.3} & 73.9 & 72.8 \\
slk & 93.5 & 90.0 & 92.2 & 70.4 & \textbf{95.7} & 71.1 & 92.4 & 95.2 & \textbf{95.7} & 94.4 & 95.2 \\
tur & 93.7 & 57.9 & \textbf{95.2} & 80.2 & 92.9 & 79.4 & 66.7 & \textbf{95.2} & 92.9 & 91.3 & \textbf{95.2} \\
vep & \textbf{71.5} & 58.8 & 70.0 & 57.5 & 68.8 & 59.2 & 60.4 & 70.7 & 68.8 & 67.8 & 70.2 \\
\midrule
\textbf{avg} & \textbf{86.3} & 68.3 & 83.9 & 78.6 & 83.8 & 80.0 & 72.2 & 85.3 & 86.2 & 85.5 & 85.9 \\
\bottomrule
\end{tabular}

\end{table*}

\subsubsection{Multilingual Model}

We use a special language ID token as part of the input sequence \cite{multi-lang-id-mt-johnson-etal-2017-googles,multi-lang-id-grapheme-phoneme-peters-etal-2017-massively} by prepending it to the sequence of morphological tags. This should help the model disambiguate identical lemma strings corresponding to different languages.

Unlike previous work on multilingual inflection which usually used the \sig{} shared task data where the train set for all languages was of the same size, we need to deal with the problem of mixing corpora of substantially different sizes. If we na\"ively pooled the training datasets, the model may be overly influenced by the high-resource languages. To address this issue, we adapt a corpus upsampling method by Wang et al. (2020) \cite{upsampling-mt-wang-etal-2020-balancing}.
For better control over sampling into training batches, we also adapt a temperature value $\multiletterM{}\in[0,1]$ (our $\multiletterM = 0.5$) to smooth the distribution of training data from datasets of different sizes, see van der Goot et al. (2021) \cite{temperature-upsampling-van-der-goot-etal-2021-massive}.

\begin{table}[t]

\centering

\caption[\sig{} 2023 comparison]{\sig{} 2023 (Lemma Disjoint): exact-match accuracy (\%), with systems AZ1-4 \cite{az-st23-kwak-etal-2023-morphological}, TÜB \cite{girrbach-2023-tu-cl}, IL1-4 \cite{illinois-canby-hockenmaier-2023-framework}, baselines \cite{st23-goldman-etal-2023-sigmorphon}.
As multilingual training wasn’t clearly defined in 2023, we highlight (*) the best systems excluding \trmultiC{} (\dag).
}
\label{tab:sig23-comparison}
\centering

\begin{tabular}{l c c ccccccccc cc}
\toprule
& \multicolumn{11}{c}{Submitted systems and official baselines} & \multicolumn{2}{c}{OURS} \\
\cmidrule(lr){1-1}\cmidrule(lr){2-12}\cmidrule(lr){13-14}

lang & {AZ3} & {AZ1} & non-neur & {AZ2} & {AZ4} & {TÜB} & neur & {IL1} & {IL2} & {IL3} & {IL4} & \trmono{} & \trmultiC{}\textsuperscript{\dag} \\
\cmidrule(lr){1-1}\cmidrule(lr){2-12}\cmidrule(lr){13-14}

afb & 34.5 & 30.8 & 30.8 & 52.7 & 52.7 & 75.8 & 80.1 & 80.7 & 82.2 & 84.1 & \textbf{84.6} & 80.3 & 79.6 \\
amh & 59.9 & 65.4 & 65.4 & 74.0 & 74.0 & 83.8 & 82.2 & 88.9 & \pb{90.6} & 88.9 & 88.6 & 88.6 & \textbf{90.8} \\
arz & 75.7 & 77.2 & 77.9 & 80.8 & 80.8 & 87.6 & \pb{89.6} & 89.2 & 88.7 & 89.1 & 88.7 & 88.5 & \textbf{89.9} \\
bel & 46.2 & 68.1 & 68.1 & 64.5 & 64.5 & 56.3 & 74.5 & 73.5 & 74.7 & 72.9 & 72.9 & \pb{75.4} & \textbf{75.9} \\
dan & 64.8 & \textbf{89.5} & \textbf{89.5} & 87.4 & 87.4 & 85.7 & 88.8 & 88.8 & \textbf{89.5} & 86.5 & 87.5 & 86.0 & 89.4 \\
deu & 59.9 & 79.8 & 79.8 & 77.9 & 77.9 & 74.5 & \pb{83.7} & 79.7 & 79.7 & 80.2 & 79.7 & 80.5 & \textbf{84.8} \\
eng & 67.0 & \pb{96.6} & \pb{96.6} & 96.2 & 96.2 & 96.0 & 95.1 & 95.6 & 95.9 & 94.6 & 95.0 & 94.5 & \textbf{97.5} \\
fin & 48.2 & 80.8 & 80.8 & 80.6 & 80.6 & 67.6 & 85.4 & 79.2 & 80.6 & 85.7 & \pb{86.1} & 84.1 & \textbf{88.0} \\
fra & 76.7 & \pb{77.7} & \pb{77.7} & 76.3 & 76.3 & 67.9 & 73.3 & 69.3 & 74.7 & 71.7 & 72.9 & 77.2 & \textbf{80.0} \\
grc & 40.4 & 52.6 & 52.6 & 54.8 & 54.8 & 36.7 & 54.0 & 48.9 & 53.7 & \pb{56.0} & \pb{56.0} & 49.7 & \textbf{61.7} \\
hun & 45.9 & 74.7 & 74.7 & 74.7 & 74.7 & 75.9 & 80.5 & 76.3 & 79.8 & 84.3 & \pb{85.0} & 84.6 & \textbf{88.9} \\
hye & 88.9 & 86.3 & 86.3 & 86.2 & 88.9 & 85.9 & 91.0 & 88.4 & 91.5 & 94.4 & 94.3 & \pb{95.1} & \textbf{98.4} \\
ita & 78.0 & 75.0 & 75.0 & 63.6 & 78.0 & 84.7 & 94.1 & 95.8 & \pb{97.2} & 92.1 & 92.2 & 95.8 & \textbf{97.8} \\
jap & 67.0 & 64.1 & 64.1 & 64.1 & 67.0 & \textbf{95.3} & 26.3 & 92.8 & 94.2 & 94.9 & 94.9 & 36.2 & 48.9 \\
kat & 71.7 & 82.0 & 82.0 & 82.1 & 82.1 & 70.5 & 84.5 & 84.1 & 84.7 & 81.3 & 82.9 & \textbf{87.1} & 85.4 \\
klr & 27.8 & 54.5 & 54.5 & 53.1 & 53.1 & 96.4 & \textbf{99.5} & 99.4 & 99.4 & 99.4 & 99.4 & 99.1 & 99.2 \\
mkd & 64.9 & 91.6 & 91.6 & 90.8 & 90.8 & 86.7 & \textbf{93.8} & 91.9 & 92.4 & 92.1 & 92.4 & 92.9 & 93.2 \\
nav & 23.7 & 35.8 & 35.8 & 41.8 & 41.8 & 53.6 & 52.1 & 54.0 & 55.1 & 55.1 & \pb{55.6} & 52.6 & \textbf{60.0} \\
rus & 66.8 & 86.0 & 86.0 & 85.6 & 85.6 & 82.1 & \pb{90.5} & 87.4 & 87.3 & 84.2 & 85.5 & 87.8 & \textbf{91.9} \\
san & 47.0 & 62.2 & 62.2 & 62.1 & 62.1 & 54.5 & 66.3 & 63.3 & \pb{69.1} & 67.7 & 65.9 & 68.6 & \textbf{73.4} \\
sme & 30.1 & 56.0 & 56.0 & 49.7 & 49.7 & 58.5 & \pb{74.8} & 69.9 & 71.8 & 67.4 & 67.3 & 73.0 & \textbf{83.1} \\
spa & 86.3 & 87.8 & 87.8 & 87.4 & 87.4 & 88.7 & 93.6 & 90.9 & 91.4 & 93.8 & 93.1 & \pb{94.9} & \textbf{95.2} \\
sqi & 73.8 & 19.3 & 83.4 & 78.1 & 78.1 & 71.5 & 85.9 & 87.6 & 88.9 & \pb{92.0} & 91.6 & 89.3 & \textbf{93.0} \\
swa & 56.2 & 60.5 & 60.5 & 65.0 & 65.0 & 94.7 & 93.7 & 93.1 & 93.1 & 96.6 & 96.6 & \pb{98.4} & \textbf{99.2} \\
tur & 28.1 & 64.6 & 64.6 & 64.6 & 64.6 & 81.8 & \pb{95.0} & 90.9 & 90.8 & 90.3 & 92.0 & 93.8 & \textbf{95.8} \\
\midrule
\textbf{avg} &  57.2 & 	68.8 &	71.4 & 71.8 & 72.6 & 76.5 & 81.1 & 82.4 & 83.9 &  83.8 & \pb{84.0}  & 82.2 &	\textbf{85.6} \\
\bottomrule
\end{tabular}

\end{table}

\begin{table}
    \centering

    \caption[UD test evaluation using \uacc{}]{UD test accuracy --- part 1. The \copybs{} baseline copies the input lemma to output. \dag The \copybs{} baseline fails on lemma-to-form transfer in Ancient\_Hebrew-PTNK due to special diacritics (e.g., cantillation marks) present in surface forms but absent in standardized lemmas.
} 
    \label{tab:test-all-langs-mono-vs-multi-uacc-p1}

\begin{tabular}{l r @{\hspace{0.3cm}} r @{\hspace{0.3cm}} r @{\hspace{0.5cm}} S[table-format=2.2] S[table-format=2.2] S[table-format=2.2]}
\toprule
& \multicolumn{3}{c}{size (\# \ltf{})} & \multicolumn{3}{c}{accuracy (\%)} \\
\cmidrule(lr){2-4}\cmidrule(lr){5-7}
lang-corpus & train & dev & test & \copybs{} & \trmono{} & \trmultiC{} \\
\midrule
Afrikaans-AfriBooms & 1.9K & 2.4K & 2.3K & 69.18 & 83.72 & \textbf{89.41} \\
Ancient\_Greek-PROIEL & 12.8K & 11.7K & 11.2K & 10.67 & 53.63 & \textbf{56.14} \\
Ancient\_Hebrew-PTNK & 4.8K & 2.8K & 2.7K & 0.00\textsuperscript{\dag} & 1.40 & \bfseries 1.66 \\
Arabic-PADT & 14.7K & 12.2K & 11.9K & 24.08 & 86.94 & \textbf{87.17} \\
Armenian-ArmTDP & 6.5K & 3.7K & 3.6K & 35.69 & 88.06 & \textbf{91.04} \\
Basque-BDT & 13.0K & 7.8K & 7.7K & 39.74 & 87.13 & \textbf{88.42} \\
Belarusian-HSE & 21.4K & 19.4K & 19.3K & 40.88 & 88.02 & \textbf{89.15} \\
Breton-KEB & 1.2K & 696 & 718 & 57.66 & 64.62 & \textbf{75.91} \\
Bulgarian-BTB & 10.0K & 9.1K & 9.0K & 37.90 & 90.94 & \textbf{92.00} \\
Catalan-AnCora & 7.4K & 15.6K & 15.4K & 64.73 & 93.62 & \textbf{95.39} \\
Chinese-GSDSimp & 7.2K & 7.7K & 7.7K & 82.10 & 80.88 & \textbf{86.45} \\
Classical\_Armenian-CAVaL & 2.6K & 2.9K & 2.6K & 28.52 & 55.64 & \textbf{72.20} \\
Classical\_Chinese-Kyoto & 3.9K & 6.4K & 6.4K & 13.99 & 1.62 & \textbf{46.27} \\
Coptic-Scriptorium & 685 & 1.5K & 1.3K & \textbf{83.18} & 54.86 & 79.45 \\
Croatian-SET & 17.7K & 12.6K & 13.0K & 36.10 & 92.10 & \textbf{93.45} \\
Czech-PDT & 47.9K & 62.0K & 62.3K & 37.53 & \textbf{97.44} & 97.37 \\
Danish-DDT & 5.8K & 6.6K & 6.7K & 55.61 & 89.92 & \textbf{92.95} \\
Dutch-Alpino & 7.0K & 11.7K & 11.8K & 55.48 & 77.66 & \textbf{79.16} \\
English-EWT & 6.6K & 9.5K & 9.7K & 76.67 & 93.63 & \textbf{94.96} \\
Erzya-JR & 4.3K & 1.7K & 1.6K & 29.63 & 82.27 & \textbf{86.10} \\
Estonian-EDT & 31.7K & 27.8K & 28.2K & 23.24 & \textbf{88.83} & 87.93 \\
Finnish-TDT & 26.3K & 15.2K & 15.0K & 23.88 & \textbf{90.58} & 88.84 \\
French-GSD & 10.4K & 20.0K & 19.6K & 72.55 & 95.36 & \textbf{96.79} \\
Galician-TreeGal & 2.4K & 1.7K & 1.7K & 59.67 & 94.09 & \textbf{97.36} \\
Georgian-GLC & 988 & 229 & 224 & 44.64 & 68.30 & \textbf{75.45} \\
German-GSD & 25.5K & 23.7K & 23.8K & 75.22 & 89.12 & \textbf{89.37} \\
Gothic-PROIEL & 4.1K & 3.1K & 2.9K & 17.50 & 71.13 & \textbf{79.30} \\
Greek-GDT & 5.2K & 4.0K & 3.9K & 34.45 & 82.69 & \textbf{86.18} \\
Hebrew-HTB & 6.7K & 6.9K & 7.0K & 51.75 & 88.78 & \textbf{90.47} \\
Hindi-HDTB & 10.4K & 12.3K & 12.1K & 79.64 & 92.31 & \textbf{92.71} \\
Hungarian-Szeged & 7.4K & 3.4K & 3.4K & 51.34 & 92.55 & \textbf{93.23} \\
Icelandic-Modern & 4.6K & 4.5K & 4.5K & 35.71 & 70.62 & \textbf{77.57} \\
Indonesian-GSD & 7.3K & 7.4K & 7.6K & 88.57 & 89.89 & \textbf{91.59} \\
Irish-IDT & 6.8K & 6.8K & 6.8K & 51.93 & 83.75 & \textbf{87.74} \\
Italian-ISDT & 7.8K & 11.9K & 11.7K & 62.95 & 95.26 & \textbf{96.02} \\
Japanese-GSDLUW & 8.4K & 11.6K & 11.6K & 74.77 & 75.31 & \textbf{79.74} \\
Korean-Kaist & 45.9K & 28.3K & 28.5K & 10.06 & 95.90 & \textbf{96.19} \\
\bottomrule
\end{tabular}
\end{table}

\begin{table}
    \centering

    \caption[UD test evaluation]{UD test accuracy --- part 2. The \copybs{} baseline copies the input lemma to output.}
    \label{tab:test-all-langs-mono-vs-multi-uacc-p2}
    
\begin{tabular}{l r @{\hspace{0.3cm}} r @{\hspace{0.3cm}} r @{\hspace{0.5cm}} S[table-format=2.2] S[table-format=2.2] S[table-format=2.2]}
\toprule
& \multicolumn{3}{c}{size (\# \ltf{})} & \multicolumn{3}{c}{accuracy (\%)} \\
\cmidrule(lr){2-4}\cmidrule(lr){5-7}
lang-corpus & train & dev & test & \copybs{} & \trmono{} & \trmultiC{} \\
\midrule
Kyrgyz-KTMU & 2.9K & 675 & 673 & 42.20 & 64.34 & \textbf{66.57} \\
Latin-ITTB & 5.8K & 9.7K & 9.7K & 14.58 & 76.33 & \textbf{88.94} \\
Latvian-LVTB & 22.6K & 18.3K & 18.2K & 27.81 & \textbf{96.48} & 96.15 \\
Lithuanian-ALKSNIS & 9.3K & 5.0K & 5.0K & 27.79 & 92.74 & \textbf{93.38} \\
Low\_Saxon-LSDC & 3.5K & 1.9K & 1.8K & 54.42 & 54.14 & \textbf{62.49} \\
Maghrebi\_Arabic\_French-Arabizi & 5.9K & 1.8K & 1.8K & \textbf{17.46} & 9.54 & 15.35 \\
Manx-Cadhan & 762 & 1.1K & 1.2K & 65.15 & 66.84 & \textbf{82.03} \\
Marathi-UFAL & 759 & 302 & 302 & 47.02 & 56.62 & \textbf{74.17} \\
Naija-NSC & 916 & 2.4K & 2.3K & 86.41 & 70.64 & \textbf{96.04} \\
North\_Sami-Giella & 4.3K & 2.1K & 2.1K & 33.00 & 68.50 & \textbf{75.96} \\
Norwegian-Bokmaal & 7.4K & 14.8K & 14.2K & 53.62 & 93.08 & \textbf{95.26} \\
Old\_Church\_Slavonic-PROIEL & 24.7K & 12.9K & 13.2K & 6.35 & 24.34 & \textbf{27.56} \\
Old\_East\_Slavic-TOROT & 28.6K & 16.7K & 16.2K & 7.49 & 31.08 & \textbf{33.82} \\
Old\_French-PROFITEROLE & 19.4K & 2.8K & 3.0K & \textbf{19.82} & 0.13 & 18.05 \\
Ottoman\_Turkish-BOUN & 3.0K & 781 & 789 & 47.66 & 78.20 & \textbf{84.03} \\
Persian-PerDT & 12.4K & 14.7K & 14.2K & 67.70 & \textbf{72.96} & 70.73 \\
Polish-PDB & 28.3K & 22.6K & 23.1K & 28.42 & \textbf{93.78} & 93.64 \\
Pomak-Philotis & 3.0K & 2.2K & 2.2K & 20.98 & 43.21 & \textbf{54.99} \\
Portuguese-Bosque & 9.7K & 11.6K & 11.7K & 67.66 & 96.21 & \textbf{97.63} \\
Romanian-RRT & 11.1K & 11.7K & 11.6K & 42.00 & 92.65 & \textbf{93.60} \\
Russian-SynTagRus & 44.4K & 64.1K & 63.2K & 26.35 & 93.28 & \textbf{93.57} \\
Sanskrit-Vedic & 19.8K & 11.3K & 11.7K & 9.43 & \textbf{75.55} & 75.22 \\
Scottish\_Gaelic-ARCOSG & 2.7K & 4.1K & 4.0K & 59.71 & 60.65 & \textbf{68.46} \\
Slovak-SNK & 14.1K & 8.1K & 7.8K & 31.83 & 94.13 & \textbf{94.45} \\
Slovenian-SSJ & 22.3K & 18.3K & 18.2K & 33.31 & 95.34 & \textbf{96.06} \\
Spanish-AnCora & 9.5K & 17.7K & 17.7K & 61.38 & 95.16 & \textbf{96.59} \\
Swedish-Talbanken & 4.9K & 5.7K & 5.6K & 48.63 & 89.63 & \textbf{92.51} \\
Tamil-TTB & 2.3K & 727 & 734 & 50.54 & 68.66 & \textbf{69.62} \\
Turkish-BOUN & 23.4K & 9.1K & 9.2K & 42.80 & 81.77 & \textbf{83.00} \\
Ukrainian-IU & 16.4K & 9.7K & 9.6K & 32.11 & 92.56 & \textbf{93.81} \\
Urdu-UDTB & 9.0K & 6.5K & 6.7K & 82.32 & 87.65 & \textbf{89.05} \\
Uyghur-UDT & 8.8K & 2.1K & 2.0K & 44.03 & 90.36 & \textbf{93.67} \\
Vietnamese-VTB & 2.2K & 2.7K & 2.7K & 92.96 & 97.74 & \textbf{99.44} \\
Welsh-CCG & 3.0K & 2.8K & 2.8K & 60.11 & 84.91 & \textbf{88.80} \\
Western\_Armenian-ArmTDP & 11.1K & 7.5K & 7.5K & 36.33 & 91.63 & \textbf{93.39} \\
Wolof-WTB & 2.3K & 2.3K & 2.4K & 78.34 & 87.35 & \textbf{89.76} \\
\midrule
\textbf{total size / avg accuracy} & 809.6K & 723.3K & 720.2K & 45.27 & 76.67 & \textbf{81.36} \\
\bottomrule
\end{tabular}

\end{table}

\section{Results}

\subsubsection{\sig{} Benchmarks}

Our monolingual and multilingual systems achieve competitive performance on a global inflection benchmark, specifically the \sig{} Inflection Shared Tasks from 2022 \cite{st22-kodner-etal-2022-sigmorphon} and 2023 \cite{st23-goldman-etal-2023-sigmorphon}, as shown in Table~\ref{tab:sig22-comparison} and Table~\ref{tab:sig23-comparison}, respectively: 
\trmono{} ranks 4th in the 2022 data and 6th in the 2023 data, while \trmultiC{} attains 3rd place in 2022 and 1st place in 2023, based on the average across languages.
For clarity, we note that the models were trained from scratch exclusively on the SIGMORPHON benchmark data.

\newcommand{\numLangsCopyWins}[0]{3}
\newcommand{\numLangsMonoWins}[0]{7}
\newcommand{\absPercentDiffMultiMono}[0]{4.69}

\subsubsection{Universal Dependencies} The two-part Table~\ref{tab:test-all-langs-mono-vs-multi-uacc-p1}~and~\ref{tab:test-all-langs-mono-vs-multi-uacc-p2} presents the results on 73 corpora from UD. The overall best-performing system is \trmultiC{}, which was trained jointly on all 73 languages. Among these languages, it is outperformed in \numLangsCopyWins{} cases by the simple \copybs{} baseline, and in \numLangsMonoWins{} cases by the monolingual models (\trmono{}), each trained separately for a single language. 
On macro average across all languages, the \trmultiC{} model improves over the \trmono{} models by \absPercentDiffMultiMono{}\%.

\section{Conclusions}

We focused on the task of automatic morphological inflection in a multilingual setting, with the aim of enabling future deployment in an open-source tool or web service. Trained jointly on 73 languages, our multilingual model is remarkably lightweight (\params{} trainable parameters), handles unseen words well, works across parts of speech, and importantly, outperforms monolingual models in most languages. We designed and implemented a novel frequency-weighted, lemma-disjoint train-dev-test split method, which combines recent evaluation practices (lemma-disjoint splits) with a realistic frequency distribution. This work establishes a foundation for deploying a practical and multilingual inflection system.

\subsubsection*{\ackname}
This research was supported by the Johannes Amos Comenius Programme (P JAC) project No. CZ.02.01.01/00/22\_008/0004605, Natural and anthropogenic georisks. 
Computational resources for this work were provided by the e-INFRA CZ project (ID:90254), supported by the Ministry of Education, Youth and Sports of the Czech Republic.
The work described herein uses resources hosted by the
LINDAT/CLARIAH-CZ Research Infrastructure (projects LM2018101 and LM2023062, supported by the Ministry of Education, Youth and Sports of the Czech Republic). 
The authors would like to thank Aleš Manuel Papáček and Milan Straka for their thoughtful comments.

\subsubsection{\discintname}
The authors have no competing interests to declare that are relevant to the content of this article.

\section{Appendix: Hyperparameters}

\newcommand{\hlay}[0]{$3/3$ (\trmono{}) and $4/4$~(\trmultiC{})}
\newcommand{\hdim}[0]{$256$}
\newcommand{\hffnn}[0]{$64$}
\newcommand{\hheads}[0]{$4$}
\newcommand{\hdrop}[0]{$0.15$}
\newcommand{\hattndrop}[0]{$0.1$}
\newcommand{\hactivdrop}[0]{$0.35$}
\newcommand{\hlayerdrop}[0]{$0.2$}
\newcommand{\hbs}[0]{$512$ (\trmono{}) and $1024$ (\trmultiC{})}
\newcommand{\hepochs}[0]{$960$}
\newcommand{\htrainparams}[0]{\paramsMono/\params{}}

For our \trmono{} and \trmultiC{} model, hyperparameters tuned on UD
development data are: layers (encoder/decoder) = \hlay{},
layer dimension = \hdim{}, feed-forward dimension = \hffnn{}, attention heads = \hheads{}, dropout = \hdrop{}, attention dropout = \hattndrop{}, activation dropout = \hactivdrop{}, and layer drop = \hlayerdrop{}. We use gradient clipping (max norm 1.0), L2 regularization (0.01), and train with Adam 
using cosine learning rate decay, initial LR~=~0.001, batch size = \hbs{}. 
Checkpoints are selected by dev set performance: on the target language for \trmono{}, or macro-averaged across languages for \trmultiC{}, over up to \hepochs{} epochs.
The \trmono{} model has \paramsMono{} parameters,\footnote{varying by vocabulary size} while the \trmultiC{} model has \params{} parameters.

\bibliographystyle{splncs04}
\bibliography{bibliography}

\end{document}